\documentclass[twocolumn]{svjour3}          
\pdfoutput=1 
\smartqed  
\usepackage{graphicx}
\usepackage{times}
\usepackage{epsfig}
\usepackage{graphicx}
\usepackage{amsmath}
\usepackage{amssymb}
\usepackage{subfigure}
\usepackage{booktabs}
\usepackage{picins}
\usepackage[table]{xcolor}
\usepackage[ruled,linesnumbered]{algorithm2e}
\usepackage{color}
\usepackage{multirow}
\usepackage[small]{caption}

 \newcommand{\CC}{C\nolinebreak\hspace{-.05em}\raisebox{.4ex}{\tiny\bf +}\nolinebreak\hspace{-.10em}\raisebox{.4ex}{\tiny\bf +}}
 
 \def\CC{{C\nolinebreak[4]\hspace{-.05em}\raisebox{.4ex}{\tiny\bf ++}}}
\begin{document}

\title{Continuous Localization and Mapping of a Pan Tilt Zoom Camera for Wide Area Tracking}


\author{Giuseppe Lisanti         \and
        Iacopo Masi \and
        Federico Pernici \and
		Alberto Del Bimbo
        }


\institute{Media Integration and Communication Center, University of Florence, \\
Viale Morgagni 65, Florence, 50134, Italy\\
+39 055 275-1390
}

\date{Received: March 18, 2015}

\maketitle

\begin{abstract}
Pan-tilt-zoom (PTZ) cameras are powerful to support object identification and recognition in far-field scenes. However, the effective use of PTZ cameras in real contexts is complicated by the fact that a continuous on-line camera calibration is needed and the absolute pan, tilt and zoom positional values provided by the camera actuators cannot be used because are not synchronized with the video stream. So, accurate calibration must be directly extracted from the visual content of the frames. Moreover, the large and abrupt scale changes, the scene background changes due to the camera operation and the need of camera motion compensation make target tracking with these cameras extremely challenging. In this paper, we present a solution that provides continuous on-line calibration of PTZ cameras which is robust to rapid camera motion, changes of the environment due to illumination or moving objects and scales beyond thousands of landmarks. The method directly derives the relationship between the position of a target in the 3D world plane and the corresponding scale and position in the 2D image, and allows real-time tracking of multiple targets with high and stable degree of accuracy even at far distances and any zooming level.
\end{abstract}

\section{Introduction}

Pan-tilt-zoom (PTZ)  cameras are powerful to support object identification and recognition in far-field scenes. They are equipped with adjustable optical zoom lenses that  can be manually or automatically controlled to permit both wide area coverage and close-up views at high resolution. This capability is particularly useful in surveillance applications to permit tracking of targets in high resolution and zooming in on biometric details of parts of the body in order to resolve ambiguities and understand target behaviors. 

However, the practical use of PTZ cameras in real contexts of operation is complicate due to several reasons. 
First, the geometrical relationship between the camera view and the 3D observed scene is time-varying and depends on camera calibration. Unfortunately, the absolute pan tilt and zoom positional values provided by the camera actuators, even when they are sufficiently precise, in most cases are not synchronized with the video stream, and, for IP cameras, a constant frame rate cannot be assumed. So, accurate calibration must be extracted from the visual content of the frames. Second, the pan tilt and zooming facility may determine large and abrupt scale changes.  This prevents the assumption of smooth camera motion.  Moreover, since the scene background is continuously changing, some adaptive representation of the scene under observation becomes necessary. All these facts have significant impact also on the possibility of having effective target detection and tracking in real-time. Due to this complexity, there is a small body of literature on tracking with PTZ cameras and most of the solutions proposed were limited to either unrealistic or simple and restricted contexts of application. 

In the following, we present a novel solution that provides continuous adaptive calibration of a PTZ camera and enables real-time tracking of targets in 3D world coordinates in general contexts of application. We demonstrate that the method is effective and is robust over long time periods of operation.

The solution has two distinct stages. In the off-line stage, we collect a finite number of keyframes taken from different viewpoints, and for each keyframe detect and store the scene landmarks and the  camera pose. In the on-line stage, we perform camera calibration by estimating the homographic transformation between the camera view and the 3D world plane at each time instant from the matching between the current view and the keyframes. Changes in the scene that have occurred over time due to illumination or objects are accounted with an adaptive representation of the scene under observation by updating the uncertainty in landmark localization. The relationship between target position in the 3D world plane and its position in the 2D image allows us to estimate the scale of target in each frame, compensate camera motion and perform accurate multi-target detection and tracking in 3D world coordinates.

\section{Related work}  
\label{Related}    

In the following, we review the research papers that are most relevant for the scope of this work. In particular, we review separately solutions for self-calibration and target tracking with moving and PTZ cameras.

\subsubsection*{PTZ camera self-calibration}

Hartley et al.~\cite{hartley1994} were the first to demonstrate the possibility of performing self-calibration of PTZ cameras based on image content. However, since calibration is performed off-line, their method cannot be applied in real-time contexts of operation. The method was improved in~\cite{agapito2001} with a global optimization of the parameters. 

Solutions for on-line self-calibration and pose estimation of moving and PTZ cameras were presented by several authors. Among them, the most notable contributions were in~\cite{Sinha,sinha:cviu06,klein07parallel,williams07iccv,civera08,Lovegrove2010,pamiptz}. 
Sinha and Pollefeys in~\cite{Sinha} used the method of~\cite{agapito2001} to obtain off-line a full mosaic of the scene. Feature matching and bundle adjustment were used to estimate the values of the intrinsic parameters for different pan and tilt angles at the lowest zooming level, and the same process is repeated until the intrinsic parameters are estimated for the full range of views and zoomings. In~\cite{sinha:cviu06} the same authors suggested that on-line control of a PTZ camera in closed loop could be obtained by matching the current frame with the full mosaic. However, their paper does not include any evidence of the claims nor provides any evaluation of the accuracy of the on-line calibration. 
Civera et al.~\cite{civera08}, proposed a method that exploits real-time sequential mosaicing of a scene. They used Simultaneous Localization and Mapping (SLAM) with Extended Kalman Filter (EKF) to estimate the location and orientation of a PTZ camera and included the landmarks of the scene in the filter state. This solution cannot scale with the number of scene landmarks. Moreover, they only considered the case of camera rotations, and did not account for zooming. 
Lovegrove et al.~\cite{Lovegrove2010} obtained the camera parameters between consecutive images by whole image alignment. As an alternative to using EKF sequential filtering, they suggested to use keyframes to achieve scalable performance. They claimed to provide full PTZ camera self-calibration but did not demonstrate calibration with variable focal length. 
The main drawback of all these methods is that they assume that the scene is almost stationary and changes are only due to camera motion, which is a condition that is unlikely to happen in real contexts.

Wu and Radke~\cite{pamiptz} presented a method for on-line PTZ camera self-calibration based on a camera model that accounts for changes of focal length and lens distortion at different zooming levels. The authors claimed robustness to smooth scene background changes and 
drift-free operation, with higher calibration accuracy than~\cite{Sinha,sinha:cviu06} especially at high zoom levels. However, as reported by the authors, this method fails when a large component in the scene abruptly modifies its position or the background changes slowly. It is therefore mostly usable with stationary scenes. A similar strategy was also applied in~\cite{Song06}, but accounts for pan and tilt camera movements, only.

Other authors developed very effective methods for pose estimation of moving cameras with pre-calibrated  internal camera parameters~\cite{klein07parallel,williams07iccv}. In~\cite{klein07parallel}, Klein and Murray applied on-line bundle adjustment to the five nearest keyframes sampled every ten frames of the sequence. In~\cite{williams07iccv}, Williams et al. used a randomized lists classifier to find  the correspondences between the features in the current view and the (pre-calculated) features from all the possible views of the scene, with RANSAC refinement. However both these approaches, if applied to a PTZ camera, are likely to produce over-fitting in the estimation of the camera parameters at progressive zoomings in.

\subsubsection*{Tracking with PTZ cameras}

Solutions to perform general object tracking with PTZ cameras were proposed by a few authors. 
Hayman et al.~\cite{Hayman99} and  Tordoff et al.~\cite{tordoff04} proposed solutions to adapt the PTZ camera focal length to compensate the changes of target size, assuming a single target  in the scene and fixed scene background.  In particular, in~\cite{Hayman99}, the authors used the affine transform applied to lines and points of the scene background; in~\cite{tordoff04} the PTZ camera focal length is adjusted to compensate depth motion of the target.
Kumar et al.~\cite{kumar02} suggested to adapt the variance of the Kalman filter to the target shape changes. They performed camera motion compensation and implemented a layered representation of spatial and temporal constraints on shape, motion and appearance. However, the method is likely to fail in the presence of abrupt scale changes.
In \cite{Varcheie:ptzTracking}, Varcheie and Bilodeau addressed target tracking with IP PTZ cameras, in the presence of low and irregular frame rate. To follow the target, they commanded the PTZ motors with the predicted target position. A fuzzy classifier is used to sample the target likelihood in each frame. Since zooming is not managed, this approach can only be applied in narrow areas. The authors in \cite{Kang:ptzTracking}  assumed that PTZ focal length is fixed and coarsely estimated from the camera CCD pixel size. They performed background subtraction by camera motion compensation to extract and track targets. This method is therefore unsuited for wide areas monitoring and highly dynamic scenes.

Solutions for tracking with PTZ cameras in specific domains of application were proposed in~\cite{kenjieccv2004,caieccv2006,seoHongIciap97,kender05}. All these methods exploit context-specific fiducial markers to obtain an absolute reference and compute the time-varying relationship between the positions of the targets in the 2D image and those in the 3D world plane.  In~\cite{caieccv2006}, the authors used the a-priori known circular shape of the hockey rink and playfield lines to locate the reference points needed to estimate the world-to-image homography and compute camera motion compensation. The hockey players were tracked using a detector specialized for hockey players trained with Adaboost and particle filtering based on the detector's confidence~\cite{kenjieccv2004}. The changes in scale of the targets was managed with simple heuristics using windows slightly larger/smaller than the current target size. 
Similar solutions were applied in soccer games~\cite{seoHongIciap97,kender05}. 

Beyond the fact that these solutions are domain-specific and have no general applicability, the main drawback is that fiducial markers are likely to be occluded and impair the quality of tracking.

\subsection{Contributions and Distinguishing Features}

The main contributions of the solution proposed are: 
\begin{itemize}   
\item We define a method for on-line PTZ camera calibration that jointly estimates the pose of the camera, the focal length and the scene landmark locations. Under reasonable assumptions, such estimation is Bayes-optimal, is very robust to zoom and camera motion and scales beyond thousands of scene landmarks. The method does not assume any temporal coherence between frames but only considers the information in the current frame.

\item We provide an adaptive representation of the scene under observation that makes PTZ camera operations independent of the changes of the scene. 

\item From the optimally estimated camera pose we infer the expected scale of a target at any image location and compute the relationship between the target position in the 2D image and the 3D world plane at each time instant.
\end{itemize} 

Differently from the other solutions published in the literature like~\cite{sinha:cviu06},~\cite{civera08},~\cite{Lovegrove2010} and~\cite{pamiptz} our approach allows performing on-line PTZ camera calibration also in dynamic scenes. Estimation of the relationship between positions in the 2D image and the 3D world plane permits more effective target detection, data association and real-time tracking. 

Some of the ideas for calibration contained in this paper were presented with preliminary results under simplified assumptions in~\cite{DLP09,DLMP10a}. Targets were detected manually in the first frame of the sequence and the scene was assumed almost static through time. Therefore we could not maintain camera calibration over hours of activity, neither support rapid camera motion. 

\begin{figure*}[t]
    \centering
	\includegraphics[width=0.85\textwidth]{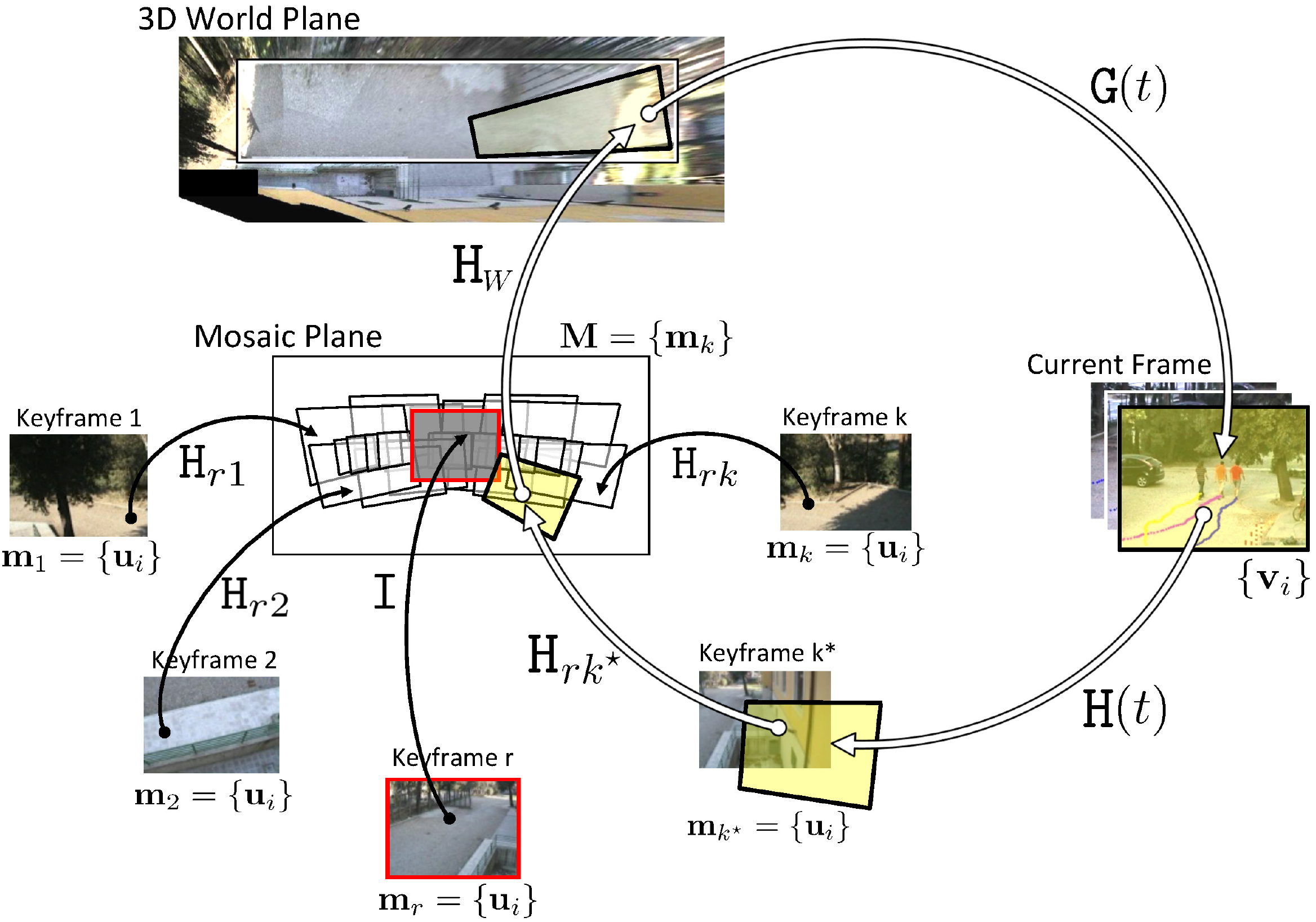}
     \caption{Main entities and their relationships: the current frame and the landmark observations extracted $\mathbf{v}$; the view maps $\mathbf{m}$ including the scene landmarks $\mathbf{u}$; the initial scene map $\mathbf{M}$ obtained from the union of the view maps; the 3D scene; the functions that represent the relationships between these entities.}
 \label{fig_fullhomo}
\end{figure*}

\section{PTZ Camera Calibration} 
\label{obsmodel}

In the following, we introduce the scene model and define the variables used. Then we discuss the off-line stage, where a scene map is obtained from the scene landmarks of the keyframes,  and the on-line stage, where we perform camera pose estimation and updating of the scene map.

\subsection{Scene model}

We consider an operating scenario where a single PTZ camera is allowed rotating around its nodal point and zooming, while observing targets that move over a planar scene. The following entities are defined as time-varying random variables:  

\begin{itemize} 
\item The \emph{camera pose } $\mathbf{c}$. Camera pose is defined in terms of the pan and tilt angles ($\psi$ and $\phi$, respectively), and focal length  $f$ of the camera. Since the principal point is a poorly conditioned parameter, it is assumed to be constant in order to obtain a more precise calibration~\cite{agapito2001}. Radial distortion was not considered since it can be assumed to be negligible for zooming operations~\cite{sinha:cviu06}.

\item The \emph{scene landmarks} $\mathbf{u}$. These landmarks account for salient points of the scene background. In the off-line stage SURF keypoints~\cite{Bay06surf:speeded} are detected in keyframe images sampled at fixed intervals of pan, tilt and focal length. A SURF descriptor is associated to each landmark. These landmarks change during the on-line camera operation.

\item The \emph{view map} $\mathbf{m}$ and \emph{scene map} $\mathbf{M}$. A view map is created for each keyframe that collects the scene landmarks (i.e. $\mathbf{m}$ = $\{\mathbf{u}_i\}$).  The scene map is obtained as the union of all the view maps and collects all the scene landmarks that have been detected at different pan, tilt and focal lengths values (i.e. $\mathbf{M}$ = $\{\mathbf{m}_k\}$). Since the scene landmarks change through time, these maps will change accordingly.  

\item The \emph{landmark observations} $\mathbf{v}$. These landmarks account for the salient points that are detected in the current frame. They can either  belong to the scene background or to targets. The SURF descriptors of the landmark observations $\mathbf{v}$ are matched with the descriptors of the scene landmarks $\mathbf{u}$, in order to estimate the camera pose and update the scene map.

\item The \emph{target state}  $\mathbf{s}$. The target state is represented in 3D world coordinates and includes both the position and speed of a target. It is assumed that targets move on a planar surface, i.e. $Z=0$, so that $\mathbf{s} = [X, Y, \dot{X}, \dot{Y}]$.

\item The \emph{target observations} in the current frame, $\mathbf{p}$. This  is a location  in the current frame that is likely to correspond to the location of a target. At each time instant $t$ there is a non-linear and time varying function $\mathbf{g}$ relating the position of the target in world coordinates $\mathbf{s}$ to the location $\mathbf{p}$ of the target in the image. Its estimation depends on the camera pose $\mathbf{c}$ and the scene map $\mathbf{M}$ at time $t$.
\end{itemize} 

Fig.~\ref{fig_fullhomo} provides an overview of  the main entities of the scene model and their relationships.

\subsection{Off-line Scene Map Initialization} 
\label{sec:scenemapinit}

In the off-line stage, image views (keyframes) are taken at regular samples of pan and tilt angles and focal length, and view maps  $\mathbf{m}_k$ are created so to cover the entire scene. SURF keypoints~\cite{Bay06surf:speeded} are organized in a k-d tree for each view map.

Given a reference keyframe and the corresponding view map $\mathbf{m}_r$, the homography that maps each $\mathbf{m}_k$ to $\mathbf{m}_r$ can be estimated as in the usual way of planar mosaicing~\cite{hartley1994}:
\begin{equation} 
\mathtt{H}_{rk}=\mathtt{K}_r\mathtt{R}_r \mathtt{R}_{k}^{-1}\mathtt{K}_{k}^{-1}
\label{eq:offline}
\end{equation}
The optimal values of both the external camera parameter matrix $\mathtt{R}_k $ and the internal camera parameter matrix $\mathtt{K}_k $ are estimated by bundle adjustment for each keyframe $k$. 

\begin{figure*}[t]
\centering
\includegraphics[width=0.9\textwidth]{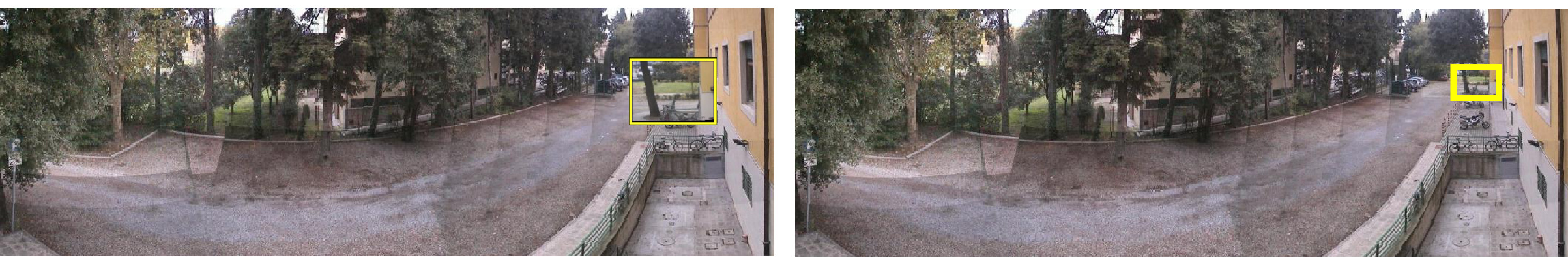}
 (a)~~~~~~~~~~~~~~~~~~~~~~~~~~~~~~~~~~~~~~~~~~~~~~~~~~~~~~~~~~~~~~~(b)
\caption{Estimations of the camera focal length of the last frame of a sequence with right panning and progressive zooming in: a) using the on-line bundle adjustment of~\cite{klein07parallel}; b) using our off-line solution with keyframes obtained by uniform sampling of the camera parameter space and the last frame.  The focal length of the last frame is represented with a square box on the scene mosaic. Focal length estimation is respectively 741.174 pixels and 2097.5 pixels. The true focal length is 2085 pixels.}
\label{fig_panos2}
\end{figure*}

Differently from~\cite{klein07parallel}, we use bundle adjustment for off-line scene map initialization and use the whole set of keyframes of the scene at multiple zoomings. Since keyframes were taken by uniform sampling of the parameter space, over-fitting of camera parameters is avoided. This results in a more accurate on-line estimation of the PTZ parameters. The difference in the accuracy of the estimation is especially sensible in the case in which PTZ operates at high zooming. Fig.~\ref{fig_panos2} shows an example of estimation of the focal length with the two approaches for a sample sequence with right panning and progressive zooming-in.

The pan, tilt, zoom values of the camera actuators are stored in order to uniquely identify each view map.  The complete scene map $\mathbf{M}$ is obtained as the union of all the view maps. Differently from~\cite{DLP09}, a forest of k-d trees is used for matching.

\subsection{On-line camera pose estimation and mapping}
\label{sec:homoest}

The positional values provided by the camera actuators at each time instant, although not directly usable for on-line camera calibration, are nevertheless sufficiently precise to retrieve the view map  $\mathbf{m}_{k^{\star}}$ with the closest values of pan, tilt and focal length. This map is likely to have almost the same content as the current frame and many landmarks will match. The landmarks matched can be used to estimate the  homography $\mathtt{H}(t)$ from the current view to $\mathbf{m}_{k^{\star}}(t)$. Matching is performed according to Nearest Neighbor distance ratio as in~\cite{lowe04} and RANSAC. To reduce the computational effort of matching, only a subset of the landmarks in $\mathbf{m}_{k^{\star}}$ is taken by random sampling. The descriptors of the landmarks matched are updated using a running average with a forgetting factor.

The optimal estimation of $\mathtt{H}(t)$ on the basis of the correspondences between landmark observations $\mathbf{v}_i (t)$ and scene landmarks  $\mathbf{u}_i (t)$ is fundamental for effective camera pose (pan, tilt, focal length) estimation and mapping in real conditions. However, changes of the visual environment due to illumination or to objects entering, leaving or changing position in the scene induce modifications of the original scene map as time progresses. Moreover, imprecisions in the detection and estimation process might affect scene landmark estimation and localization.  
To this end, under reasonable assumptions, we derive a linear measurement model that accounts for all the sources of error of landmark observations, that permits to obtain the optimal localization of the scene landmarks. Permanent modifications of the scene are accounted through a landmark birth-death process that includes new landmarks and discards temporary changes.

\subsubsection*{Closed-form recursive estimation of scene landmarks}

Camera pose estimation and mapping requires inference of the joint probability of the camera pose  $\mathbf{c}(t)$ and scene landmark locations in the map $\mathbf{M}(t)$, given the landmark observations $\mathbf{v} $ until time $t$ and the initial scene map $\mathbf{M}(0)$: 
\begin{equation} 
p \big (\mathbf{c}(t), \mathbf{M}(t) | \mathbf{v} (0:t), \mathbf{M}(0) \big ). 
\label{eq_slam} 
\end{equation} 
In order to make the problem scalable with respect to the number of landmarks, Eq.~(\ref{eq_slam}) is approximated by decoupling camera pose estimation from map updating: 
\begin{equation}
       \underbrace{p \big (\mathbf{c}(t) | \mathbf{v} (t), \mathbf{M}(t-1) \big )}_{\mbox{camera pose estimation}}  \underbrace{ \ p \big (\mathbf{M}(t) | \mathbf{v} (t), \mathbf{c}(t), \mathbf{M}(t-1)\big )}_{\mbox{map updating}}
         \label{eq_slam_decoupled}
\end{equation} 
    
Considering the the view map  $\mathbf{m}_{k^{\star}}$ with the closest values of pan, tilt and focal length and applying Bayes theorem to the map updating term, Eq.~(\ref{eq_slam_decoupled}) can be rewritten as:
\begin{eqnarray}
\nonumber
p \big (\mathbf{m}_{k^\star}(t) | \mathbf{v} (t), \mathbf{c}(t) , \mathbf{m}_{k^\star}(t-1) \big ) = \\
p \big (\mathbf{v}(t)| \mathbf{c}(t),
\mathbf{m}_{k^\star}(t) \big ) p \big (\mathbf{m}_{k^\star}(t)|\mathbf{m}_{k^\star}(t-1)  \big ),
\label{new_eq_map_update_bayes} 
\end{eqnarray}
where the term $p \big (\mathbf{m}_{k^\star}(t)|\mathbf{m}_{k^\star}(t-1)  \big)$ indicates that view map $\mathbf{m}_{k^\star}(t)$ at time $t$ depend only on $\mathbf{m}_{k^\star}(t-1)$. Assuming that for each camera pose the observation landmarks $\mathbf{v}_{i}$ that match the scene landmarks $\mathbf{u}_{i}$ in $\mathbf{m}_{k^\star}(t)$ are independent of each other, i.e.:
\begin{equation} 
p \big (\mathbf{v} (t) | \mathbf{c}(t) , \mathbf{m}_{k^\star}(t)  \big ) = \prod_i p \big (\mathbf{v} _{i}(t)| \mathbf{c}(t), \mathbf{u}_{i}(t) \big ),
\label{likelihoodmap}
\end{equation} 
Eq.~(\ref{new_eq_map_update_bayes}) modifies in:
\begin{eqnarray}
\nonumber
p \big (\mathbf{m}_{k^\star}(t) | \mathbf{v} (t), \mathbf{c}(t) , \mathbf{m}_{k^\star}(t-1) \big ) = \\
\prod_i p \big (\mathbf{v} _{i}(t)| \mathbf{c}(t),\mathbf{u}_{i}(t) \big ) p \big (\mathbf{u}_i(t)| \mathbf{u}_i(t-1)  \big ),
\label{eq_map_update} 
\end{eqnarray} 
where $p \big (\mathbf{u}_i(t)| \mathbf{u}_i(t-1)  \big )$ is the prior pdf of the $i$-th scene landmark at time $t$ given its state at time $t-1$. Under the assumptions that both scene landmarks $\mathbf{u}_{i}(t)$ and the keypoint localization error have a Gaussian distribution, and that Direct Linear Transform is used, the observation model $p \big (\mathbf{v} _{i}(t)|\mathbf{c}(t), \mathbf{u}_{i}(t) \big )$ can be expressed as:
\begin{equation}
\mathbf{v}_i(t) = \mathbf{H}_i(t)\mathbf{u}_i(t) + \mathbf{\mbox{\boldmath$\lambda$}}_{i}(t),
\label{linear_model} 
\end{equation}
where $\mathbf{H}_i(t)$  is the $2 \times 2$ matrix obtained by linearizing the homography $\mathtt{H}(t)$ at $\mathbf{v}_i(t)$ and $\mathbf{\mbox{\boldmath$\lambda$}}_{i}(t)$ is an additive Gaussian noise term with covariance $\mathbf{\Lambda}_i(t)$ that represents the whole error in the landmark mapping process. This covariance can be expressed in closed form and in homogeneous coordinates as: 
\begin{eqnarray} 
\mathtt{\Lambda}_{i}(t) = { \mathtt{B}_{i}(t) \, \mathtt{\Sigma}_{i}(t) \mathtt{B}_{i}(t)^\top + 
                       \mathtt{\Lambda}_{i}^\prime }~+ \mathtt{H}(t)^{-1} \, \mathtt{P}_{i}(t) \mathtt{H}(t)^{-\top},
\label{eq:eq_cov} 
\end{eqnarray} 
where\ the three terms account respectively for the spatial distribution of the matched landmarks, the covariance of keypoint localization in the current frame and  the uncertainty associated to the scene landmark positions in the view map. In Eq.~(\ref{eq:eq_cov}), $\mathtt{\Sigma}_{i}(t)$ is the $9 \times 9$ homography covariance matrix (calculated in closed form according to~\cite{Criminisi99a}) and $\mathtt{B}_{i}(t)$ is the $3\times 9$ block matrix  of landmark observations; $\mathtt{\Lambda}^\prime_{i}$ models the keypoint detection error covariance;   $\mathtt{P}_{i}(t)$ is the covariance of the estimated landmark position on the nearest view map, and $\mathtt{H}$ is obtained from the Direct Linear Transform. Covariance $\mathbf{\Lambda}_i(t)$ can be directly obtained as the $2\times 2$ principal minor of $\mathtt{\Lambda}_{i}(t)$.  

The optimal localization of the scene landmarks is therefore obtained in closed form through multiple applications of the Extended Kalman Filter to each landmark observation, with the Kalman gain being computed as:

\begin{equation}
\resizebox{.99\hsize}{!}{$
\mathbf{K}_i(t)=
 \mathbf{P}_i(t|t-1) {\mathbf{H}}_i(t)^{-1} \big [{\mathbf{H}}_i(t)^{-1}\mathbf{P}_i(t|t-1) {\mathbf{H}}_i(t)^{-\top} + {\mathbf{\Lambda}}_i(t) \big ]^{-1}, 
 $}
\end{equation}
where $\mathbf{P}_i$ is the Kalman covariance of the $i$-th scene landmark. 

\subsubsection*{Birth-death of scene landmarks}
\label{sec:birthdeath}

\begin{figure}[t] 
	\centering 
		\includegraphics[width=\columnwidth]{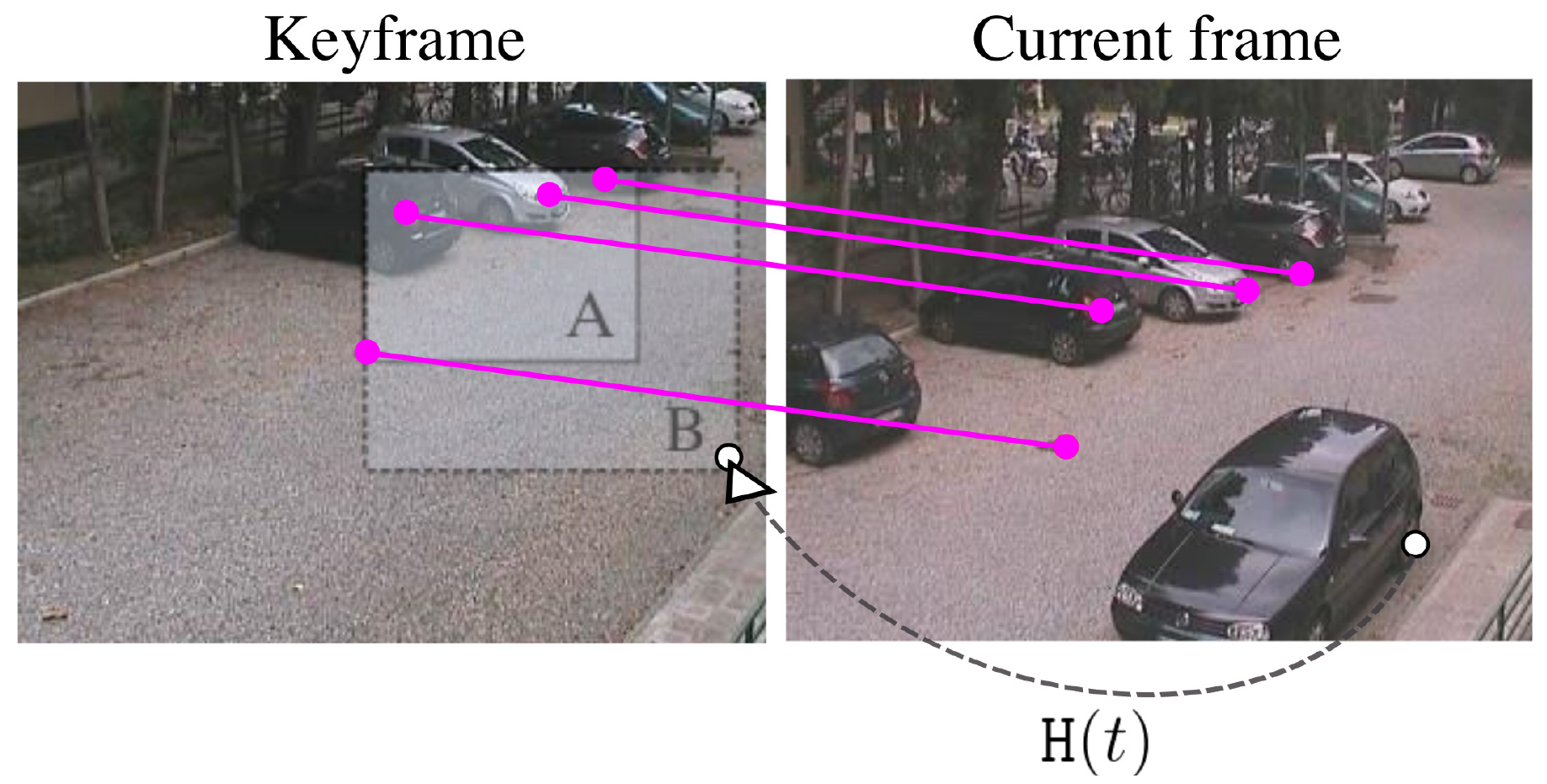} 
		\caption{\emph{Proximity check} for scene map updating. Current frame and its nearest keyframe in the scene map. Matched landmarks and a new landmark are shown in magenta and white, respectively, together with their bounding boxes.
}
\label{fig_landINOUT}
\end{figure}

Objects that enter or leave the scene introduce modifications of the original scene map. 
Their landmarks are not taken into account in the computation of $\mathtt{H}(t)$ at the current time (they are the RANSAC outliers in the matching process), but are taken into account in the long term, in order to avoid that the representation of the original scene becomes drastically different from that of the current scene. 
We assume that new landmarks that persist in 20 consecutive frames and are closest to the already matched landmarks have higher probability of belonging to a new scene element (they have smaller covariance according to Eq.~(\ref{eq:eq_cov})). According to this, we implemented a \emph{proximity check} (Fig.~\ref{fig_landINOUT}) that computes such probability as the ratio between the bounding box of the landmarks matched and the extended bounding box of the new landmark (respectively box A and B in Fig.~\ref{fig_landINOUT}).
Such candidate landmarks are included in $\mathbf{m}_{k^{\star}}$ using the homography $\mathtt{H}(t)$. Landmarks are terminated when they are no more matched in consecutive frames.

Since the transformation between two near frames under pan tilt and zoom can be locally approximated by a similarity transformation, the asymptotic stability of the updating procedure is guaranteed by the Multiplicative Ergodic Theorem~\cite{alien}. Therefore, we can assume that no sensible drifting is introduced in the scene landmark updating.

\subsubsection*{Localization in world coordinates}

\begin{figure*}[t] 
	\centering 
		\includegraphics[width=0.9\textwidth]{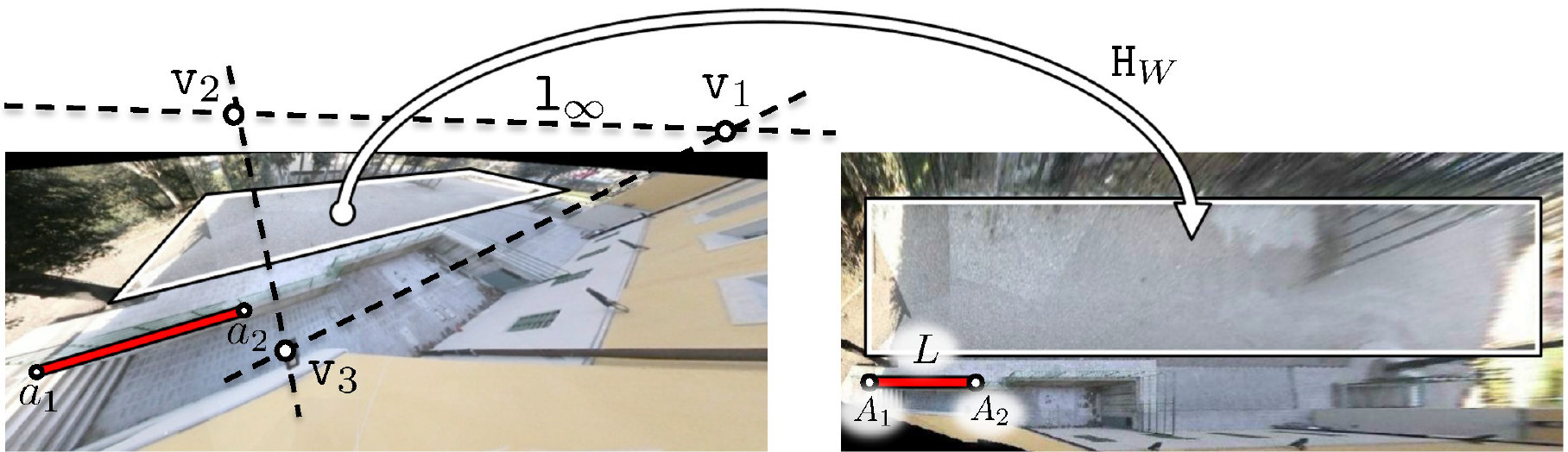} 
		\caption{The transformation from the 2D mosaic plane  (\emph{Left}) to the 3D world plane (\emph{Right}). The vanishing points and the vanishing lines are used for the computation of matrix $\mathtt{H}_p$. A pair of corresponding points to compute $\mathtt{H}_s$ is shown.} 
		\label{fig_mosaic2world} 
\end{figure*}

Looking at Fig.~\ref{fig_fullhomo}, the time varying homography $\mathtt{G}(t)$ (in homogeneous coordinates), mapping a target position in the world plane to its position $\mathbf{p} $ in the current frame, can be represented as: 
\begin{equation} 
\mathtt{G}(t) =  \big ( \mathtt{H}_W \mathtt{H}_{rk^{\star}}  \mathtt{H}(t) \big )^{-1}, 
\label{eq_homo_mosa_image} 
\end{equation} 
where $\mathtt{H}_W$ is the stationary homography from the mosaic plane to the 3D world plane: 
\begin{equation} 
\mathtt{H}_W=\mathtt{H}_s\mathtt{H}_p , 
\end{equation}
that can be obtained as the product of the rectifying homography $\mathtt{H}_p$ (derived from the projections of the vanishing points by exploiting the single view geometry of the planar mosaic\footnote{In the case of a PTZ sensor, the homography between each keyframe and the reference keyframe is the infinite homography $\mathtt{H}_\infty$ that puts in relation vanishing lines and vanishing points between the images.}~\cite{Liebowitz98})  and transformation $\mathtt{H}_s$ from pixels in the mosaic plane to 3D world coordinates (estimated from the projection of two points at a known distance $L$ in the world plane onto two points in the mosaic plane as in Fig.~\ref{fig_mosaic2world}). 

\section{Target tracking with PTZ cameras} 
\label{sec:mtt}

We perform  multi-target tracking in 3D world coordinates using the Extended Kalman Filter. Data association to discriminate between target trajectories is implemented according to the Cheap-JPDAF model~\cite{Fitzgerald1985}. 

The relationship between the image plane and the 3D world plane of Eq.~(\ref{eq_homo_mosa_image})  
allows us to obtain the target scale and perform tracking in the 3D world plane. As it will be shown in Section \ref{sec:expsection}, tracking in the 3D world plane allows a better discrimination between targets.

\subsection{Target scale estimation}%
\label{sec:detector}

At each time instant $t$, the homography $\mathtt{G}(t)$ permits to derive the homology relationship that directly provides the scale at which the target is observed in the current frame:
\begin{equation} 
\mathbf{h}(t) = \mathtt{W}(t) \mathbf{p}(t)  
        \label{eq_planHomo}  
\end{equation} 
where $\mathbf{h}$(t) and $\mathbf{p}$(t) are respectively the position of the target top and bottom in the image plane and $\mathtt{W}(t)$ is defined as: 
\begin{equation}  
        \mathtt{W}(t) =  
            \mathtt{I} +(\mu - 1)  
                \frac{ \mathbf{v}_{\infty}(t) \cdot \mathbf{l}_{\infty}^\top(t)}  
                    { \mathbf{v}_{\infty}^\top(t) \cdot  \mathbf{l}_{\infty}(t)} ,  
\end{equation}  
where $\mathtt{I}$ is the identity matrix, $\mathbf{l}_{\infty}(t)$ is the world plane vanishing line, $\mathbf{v}_{\infty}(t)$ is the vanishing point of the world normal plane direction, and $\mu$ is the cross-ratio. The vanishing point $\mathbf{v}_{\infty}(t)$ is computed  as $\mathbf{v}_{\infty}(t)  = \mathtt{K}(t) \mathtt{K}(t)^\top \cdot \mathbf{l}_{\infty}(t)$, with $\mathbf{l}_{\infty}(t)  = \mathtt{G}(t) \cdot  [0, \, 0, \, 1]^\top$ and $\mathtt{K}(t)$ is derived from $\mathtt{H}(t)$ as in~\cite{DLP09}. Estimation of the target scale allows us to apply the detector at a single scale instead of multiple scales and improve in both recall and computational performance for detection and tracking.

\subsection{Multiple Target Tracking}

The Extended Kalman filter observation model for each target is defined as:
\begin{equation}
 \mathbf{p}(t) = \mathbf{g}\big(s(t),t \big) =
\left[
  \begin{array}{cc}
    \mathbf{G}(t) & \mathbf{0}_{2 \times 2} \\
  \end{array}
\right]
\mathbf{s}(t) + \zeta(t),
\label{eq_obs_model_target}
\end{equation}
where $\zeta(t)$ is a Gaussian noise term with zero mean and diagonal covariance that models the target localization error in the current frame;  $\mathbf{s}(t)$ is the target state, represented in 3D world coordinates, $\mathbf{G}(t)$ is the homography $\mathtt{G}(t)$ linearized at the predicted target position and $\mathbf{0}_{2 \times 2}$ is the $2 \times 2$ zero matrix. Assuming constant velocity, the motion model in the 3D world plane is defined as:
\begin{equation}
    p(\mathbf{s} (t) |\mathbf{s} (t-1)) = \mathcal{N}(\mathbf{s} (t); \mathbf{A} \mathbf{s}(t-1), \mathbf{Q}),
    \label{eq_linDyn}
\end{equation}
where $\mathbf{A}$ is the $4 \times 4$ constant velocity transition matrix and $\mathbf{Q}$ is the $4 \times 4$ process noise matrix. 
For multiple target tracking, $\mathbf{{G}}(t)$ influences the target covariance of the Cheap-JPDAF respectively for the Kalman gain expression: 
\begin{equation} 
	\mathbf{W}(t) = \mathbf{P}(t|t-1) \mathbf{{G}}(t) \mathbf{S}(t|t)^{-1},
\end{equation}
and the target covariance on the image plane:
\begin{eqnarray}
{\mathbf{S}}(t|t)  =  \mathbf {{G}}(t) \mathbf{P}(t|t-1) \mathbf{{G}}(t)^{\top}   + \mathbf{V}(t), 
\end{eqnarray}
where $\mathbf{V}(t)$ is the covariance matrix of the measurement error of Eq.~(\ref{eq_obs_model_target}).

\section{Experimental results}
\label{sec:expsection}
In this Section we report on an extensive set of experiments
to assess the accuracy of our PTZ camera calibration method and its effective exploitation for real-time multiple target tracking.

\subsection{PTZ camera calibration}
In the following, we summarize the experiments that validate our approach for camera calibration. We justify the use of motor actuators to retrieve the closest scene map; we report on the precision of the off-line scene map initialization and the on-line camera pose estimation and mapping.

\subsubsection*{Accuracy of PTZ motor actuators}

\begin{figure*}
\centering
\includegraphics[width=.95\textwidth]{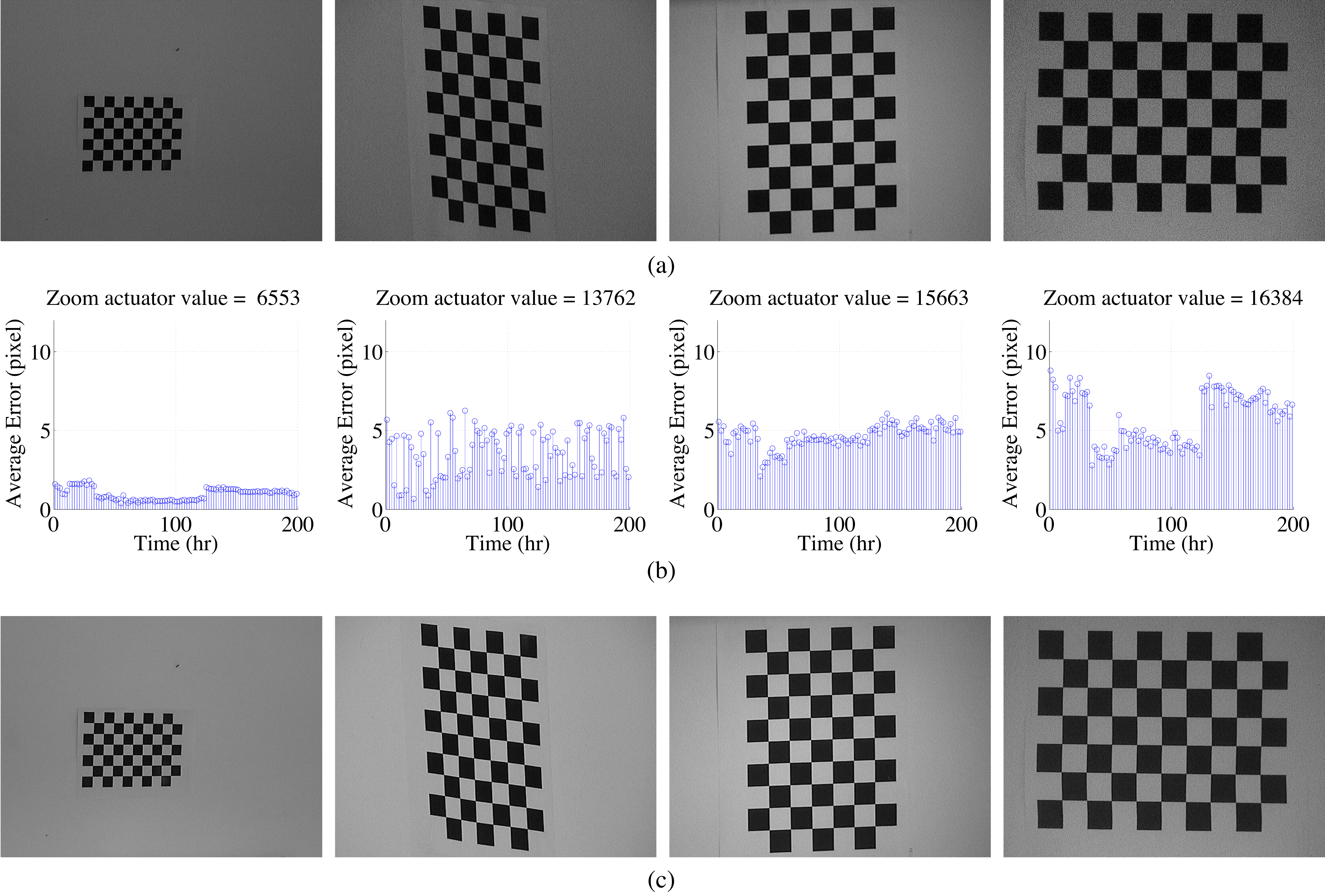}
	\caption{(a) Checkerboard images at the initial camera pose. (b) Average Errors over 200 hours. (c) Checkerboard images after the camera has returned in the same initial pose after 200 hours.}
\label{fig:ptzAct}
\end{figure*}

We validated the use of pan tilt and zoom values provided by the camera motor actuators to retrieve the closest view map, by checking their precision with the same experiment as in~\cite{pamiptz}. We placed four checkerboard targets at different positions in a room. These positions corresponded to different pan, tilt and zoom conditions. A SONY SNC-RZ30P PTZ camera was moved to a random position every 30 seconds and returned at the  initial positions every hour. For each image view the corners of the checkerboard were extracted and compared to the reference image. The errors were collected for 200 hours. We have measured an average error of 2 pixels at the lowest zooming and 9 pixels for the maximum zooming. Fig.~\ref{fig:ptzAct} shows the plots of the errors and the initial and final camera view for each target.

\subsubsection*{Scene map initialization}
\begin{figure}
\centering
\subfigure[]{\includegraphics[width=.49\columnwidth]{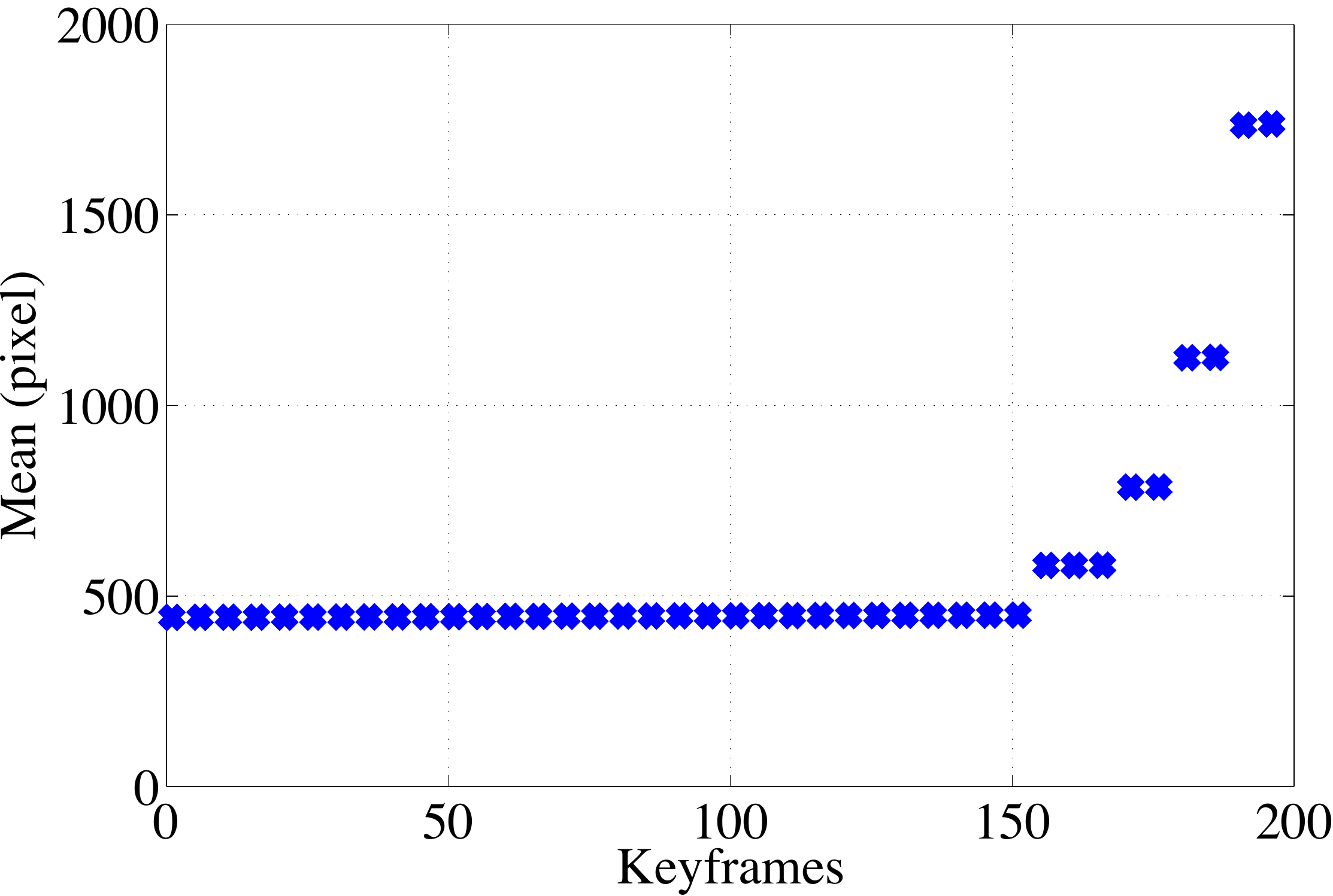}}
\subfigure[]{\includegraphics[width=.47\columnwidth]{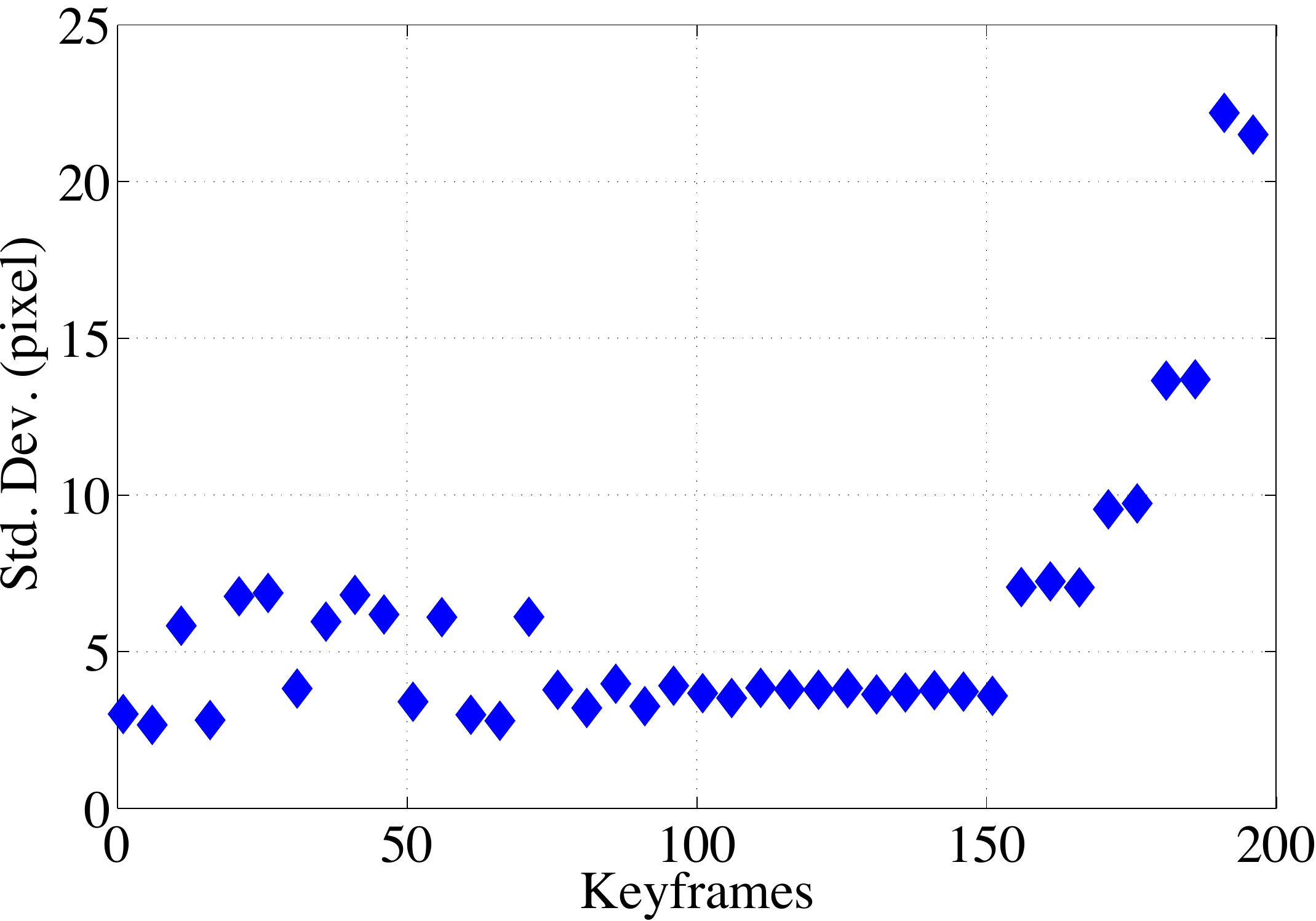}}
\caption{Average (a) and standard deviation (b) of the bundle-adjusted focal length for the keyframes used in scene map initialization. Keyframes are ordered for increasing values of focal lenght.}
\label{fig:offlineDev}
\end{figure}

Off-line scene map initialization as discussed in Sect.~\ref{sec:scenemapinit} is accurate and produces repeatable results. Fig.~\ref{fig:offlineDev} reports the mean and standard deviation of the focal length estimated during the scene map initialization.
In this experiment, we acquired images of the same outdoor scene in 43 consecutive days at different time of the day, at 202 distinct values of pan tilt zoom. The PTZ camera was driven using motor actuators. We can notice that the standard deviation of the focal length that is estimated through off-line bundle adjustment increases almost proportionally with focal length. The maximum standard deviation value observed is 23 pixels at focal length of about 1700 pixels.

\begin{table*}
\centering
\resizebox{0.9\textwidth}{!}{
\setlength{\tabcolsep}{0.6em}
\begin{tabular}{|l|c|c|c|c|c|c|c|c|c|}
\hline 	 
\multicolumn{1}{|l}{\textbf{Sequence}} & \multicolumn{1}{|c}{\textbf{\#measurements}} & \multicolumn{2}{|c}{\textbf{Avg. reproj. error}} & \multicolumn{2}{|c}{\textbf{Pan}} & \multicolumn{2}{|c}{\textbf{Tilt}} & \multicolumn{2}{|c|}{\textbf{Focal Length}}  \\  \cline{1-10}
 -- & -- & Ours &  Ours w/o p. & Ours & Ours w/o p. & Ours & Ours w/o p. & Ours & Ours w/o p. \\ \cline{1-10}
Seq.~1 & 34,209 & \textbf{2.83} & 2.96 & \textbf{1.18} & 1.55 & \textbf{0.39} & 0.42 & \textbf{0.96} & 1.06 \\ 
Seq.~2 & 34,605 & \textbf{6.69} & 6.90 & 2.47 & \textbf{2.09} & \textbf{0.68} & 0.94 & 4.41 & \textbf{3.65}  \\ 
Seq.~3 & 33,102 & \textbf{3.26} & 3.30 & 1.26 & \textbf{1.17} & 0.33 & 0.33 & \textbf{0.84} & 0.91  \\ 
Seq.~4 & 33,939 & \textbf{6.88} & 7.09 & \textbf{2.11} & 2.58 & 1.93 & \textbf{1.73} & \textbf{2.78} & 3.79 \\ 
Seq.~5 & 33,974 & \textbf{22.54} & 60.04 & \textbf{11.14} & 11.53 & \textbf{9.51} & 9.85 & \textbf{12.49} & 14.21 \\ 
Seq.~6 & 33,570 & \textbf{3.21} & 4.26 & \textbf{1.91} & 2.84 & \textbf{0.49} & 0.54 & \textbf{1.26} & 3.05 \\ 
Seq.~7 & 34,157 & 3.62 & \textbf{3.59} & 1.71 & \textbf{1.27} & \textbf{0.35} & 0.43 & \textbf{1.81} & 2.15  \\ 
Seq.~8 & 33,932 & \textbf{21.76} & 21.99 & \textbf{7.08} & 7.41 & 10.07 & \textbf{9.23} & 11.91 & \textbf{11.81} \\ 
Seq.~9 & 34,558 & \textbf{8.78} & 12.26 & \textbf{3.35} & 5.48 & \textbf{1.37} & 2.70 & \textbf{3.47} & 4.80 \\ 
Seq.~10 & 34,405 & \textbf{8.47} & 9.26  & 7.20 & \textbf{5.71} & \textbf{5.28} & 6.59 & \textbf{8.99} & 9.54 \\ \cline{1-10}
Average & 34,032 & \textbf{8.80} & 13.17 & \textbf{3.94} & 4.16 & \textbf{3.04} & 3.28 & \textbf{4.89} & 5.50 \\ \cline{1-10}
\end{tabular}
}
\caption{Average reprojection error and calibration errors of pan, tilt and focal length with and without \emph{proximity check} evaluated at the keyframes during the period of observation.}
\label{table:errCalib}
\end{table*}

\subsubsection*{On-line PTZ camera pose estimation and mapping}
\label{sec_longterm}
In this experiment, we report on the average reprojection error and calibration errors with our method. We discuss the influence of the number of landmarks and RANSAC inlier threshold on the reprojection error and the effectiveness of scene landmark updating.

As in~\cite{pamiptz}, we recorded 10 outdoor video sequences of 8 hours each (80 hours in total). Due to the long period of observation, all the sequences include slow background changes due to shadows or illumination variations, as well as large changes due to moving objects entering or exiting the scene. The PTZ camera was moved continuously using the motor actuators and stopped for a few seconds at the same pan tilt zoom values, so to have a large number of keyframes at the same scene locations and different conditions, in all the sequences. On average we performed about $34000$ measurements per sequence. For each keyframe, a grid of points was superimposed and the average reprojection error was measured between the grid points as obtained by the estimated homography and the same points by the off-line bundle adjustment.

Tab.~\ref{table:errCalib} shows the average reprojection error, the errors in the estimation of pan, tilt and focal length and the improvements that are obtained with the \emph{proximity checking}, for the outdoor sequences under test. As in~\cite{pamiptz}, the errors in pan and tilt angles were computed as $e_{\psi}(t)=\left| \psi(t) - \psi_{rk} \right |$ and $e_{\phi}(t)=\left| \phi(t) - \phi_{rk} \right|$, respectively, and  the focal length error as $e_{f}(t)=\Big| \frac{ f(t) - f_{rk}}{f_{rk}} \Big|$ (in percentage). Pan and tilt angles estimated and those calculated with bundle adjustment were obtained from the rotation matrices $\mathtt{R}(t) = \mathtt{K}_{r}^{-1} \mathtt{H}_{rk^{\star}}\mathtt{H}(t)\mathtt{K}_{k}$ (see Eq.~(\ref{eq_homo_mosa_image}))  and $\mathtt{R}_{rk} = \mathtt{K}_{r}^{-1} \mathtt{H}_{rk}\mathtt{K}_{k}$ (see Eq.~(\ref{eq:offline})), respectively.
The results confirm that \emph{proximity checking} avoids to select landmarks that introduce drifting in the homography estimation. 
It can be observed that errors in focal length measured with our method over a long period in an outdoor scenario are similar to those obtained in~\cite{pamiptz}, and lower than  those in~\cite{sinha:cviu06} (as reported in~\cite{pamiptz}), for an indoor experiment with a few keyframes.

\begin{figure}
\centering
\subfigure[]{\includegraphics[width=.485\columnwidth]{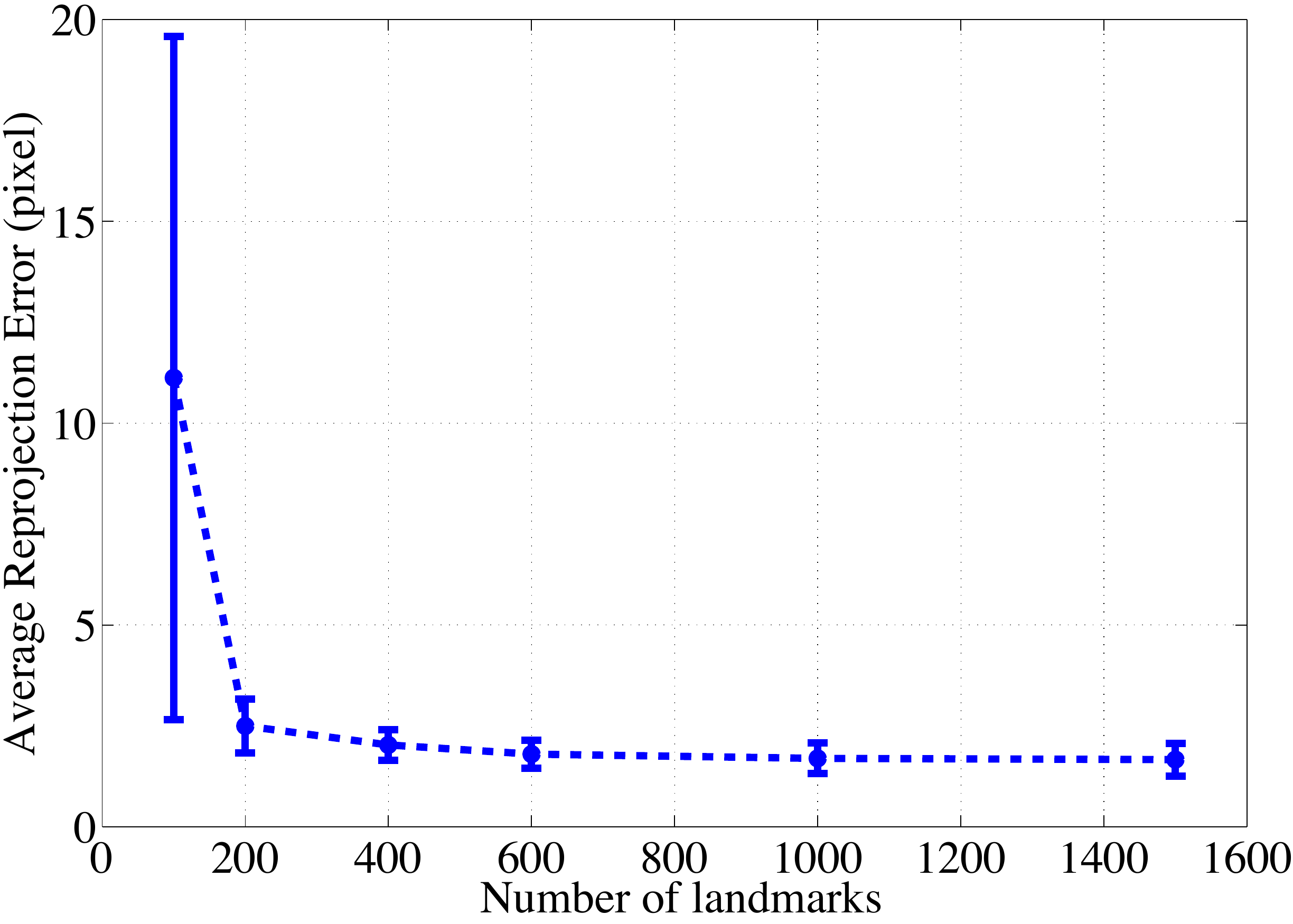}} 
\subfigure[]{\includegraphics[width=.48\columnwidth]{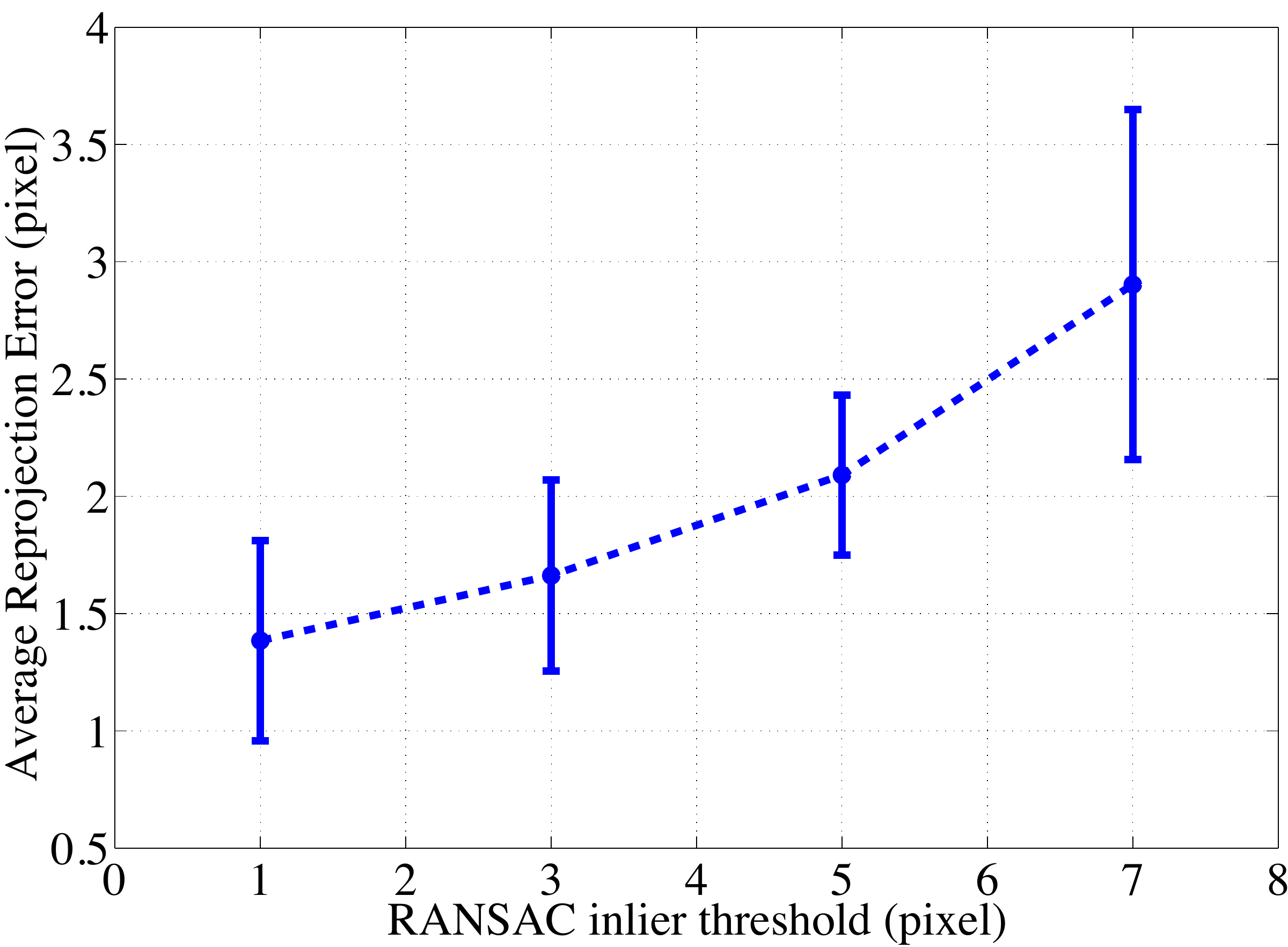}}
\caption{Reprojection error as a function of (a) the number of landmarks extracted (b) inlier threshold in the RANSAC algorithm, for Sequence 1 under test.}
\label{fig:param}
\end{figure}

The reprojection error depends on both the number of landmarks extracted and the RANSAC threshold for inliers as shown in Fig.~\ref{fig:param} for one of the sequences under test (Sequence 1). It can be observed that a large reprojection error with high standard deviation (plotted at one sigma) is present below 200 landmarks. Instead, such error is low when the number of landmarks is between 200 and 1500 (Fig.~\ref{fig:param}(a)). Fig.~\ref{fig:param}(b) shows that a RANSAC thresholds between 1 and 3 pixels for the inliers used in the homography estimation assures small reprojection errors. Values of 1000 and 3 pixels were used respectively for the number of landmarks extracted and RANSAC threshold in our experiments.

\begin{figure}
\centering
\subfigure[]{\includegraphics[width=.48\columnwidth]{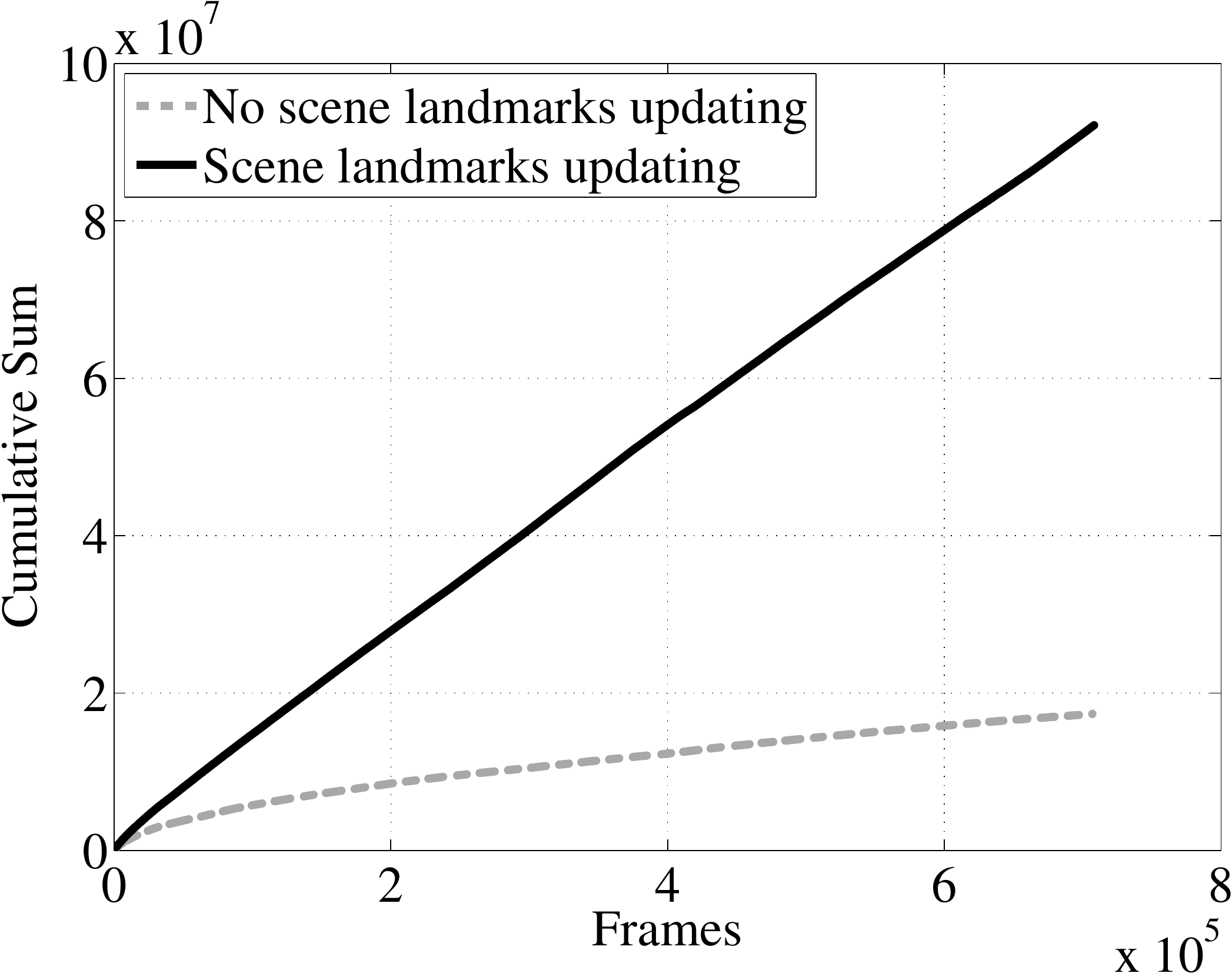}}
\subfigure[]{\includegraphics[width=.5\columnwidth]{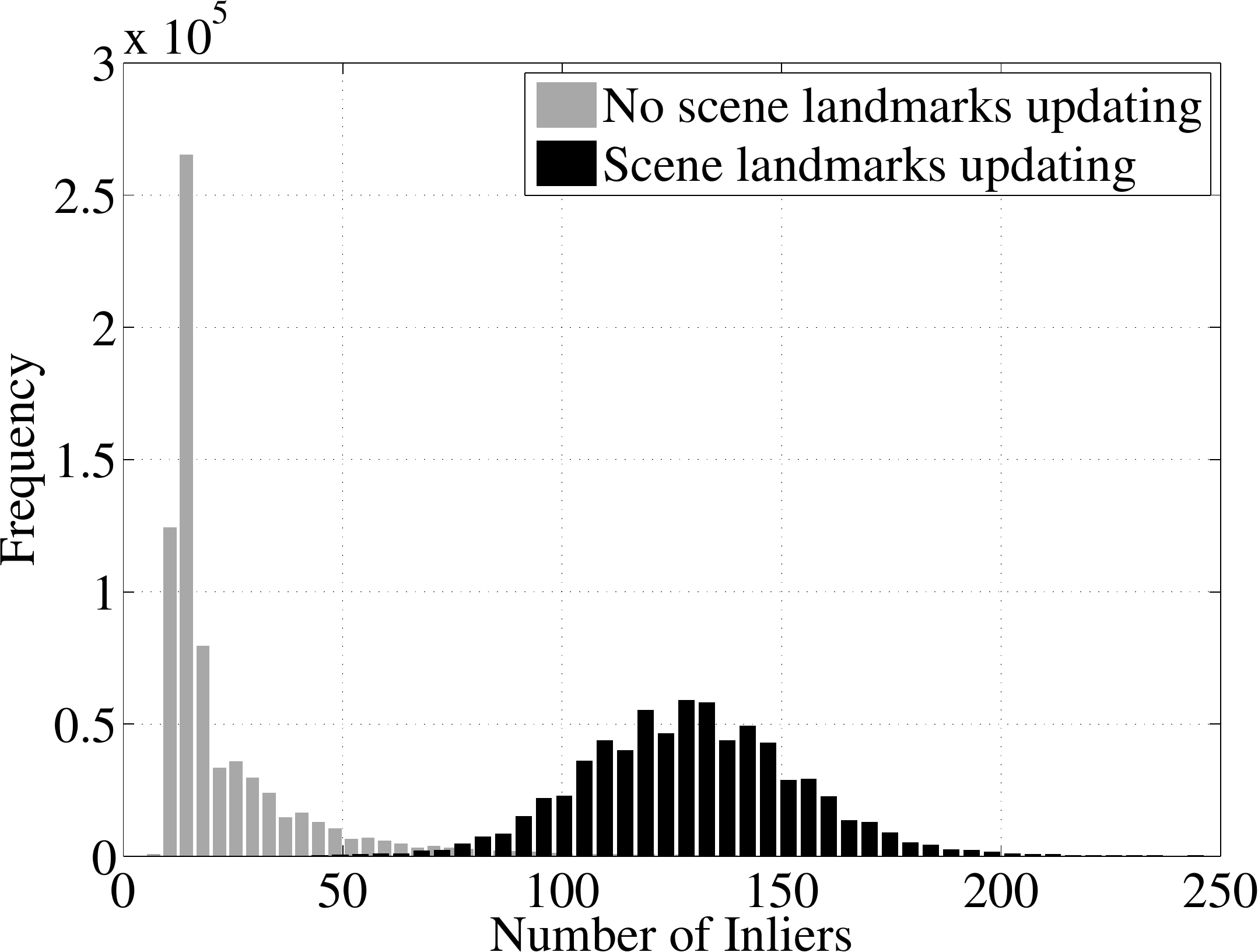}}
\caption{(a) Cumulative Sum of number of inliers as a function of time: without and with scene landmark updating (dashed and solid curve respectively). (b) Distributions of the number of inliers without and with scene landmark updating (grey and black bins respectively).}
\label{fig:inliers}
\end{figure}

\begin{figure}
\centering
\includegraphics[width=\columnwidth]{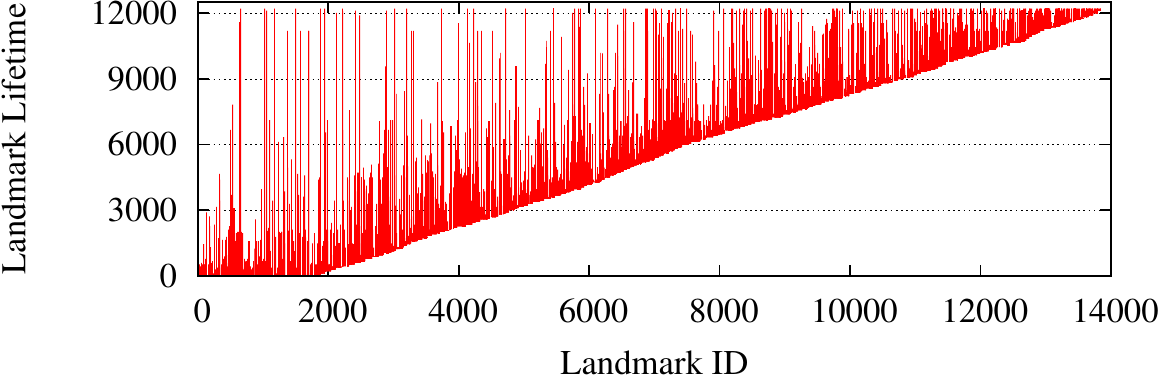}
\caption{Lifetime of scene landmarks observed for a sample keyframe.}
\label{fi:stimageometry}
\end{figure}

Scene map updating significantly contributes to the robustness of our camera calibration to both slow and sudden variations of the scene, maintaining a high number of RANSAC inliers through time.    
Fig.~\ref{fig:inliers}(a) shows the cumulative sum of the inliers with and without scene landmark updating. It is possible to observe that without scene landmark updating the number of inliers decreases (the cumulative curve is almost flat) as the initial landmarks do not match anymore with the landmarks observed due to scene changes. 
Fig.~\ref{fig:inliers}(b) shows the distribution of the inliers in the two cases. With no scene landmark updating, typically only few of the original landmarks are taken as inliers for each keyframe, that is insufficient to assure a robust calibration over time. With scene landmark updating, a higher number of inliers is taken for each frame that include both the original and the new scene landmarks. As can be inferred from Fig.~\ref{fig:inliers}, in a dynamic scene few of the original scene landmarks survive at the end of the observation period. Fig.~\ref{fi:stimageometry} highlights the scene landmark lifetime over a 20 minutes window, for one keyframe (randomly chosen). The scene landmarks with ID $\in [0..2000]$ are the original landmarks. Landmarks with ID $\geq 2000$ are those observed during the 20 minutes. 

\begin{figure}
\centering
\includegraphics[width=\columnwidth]{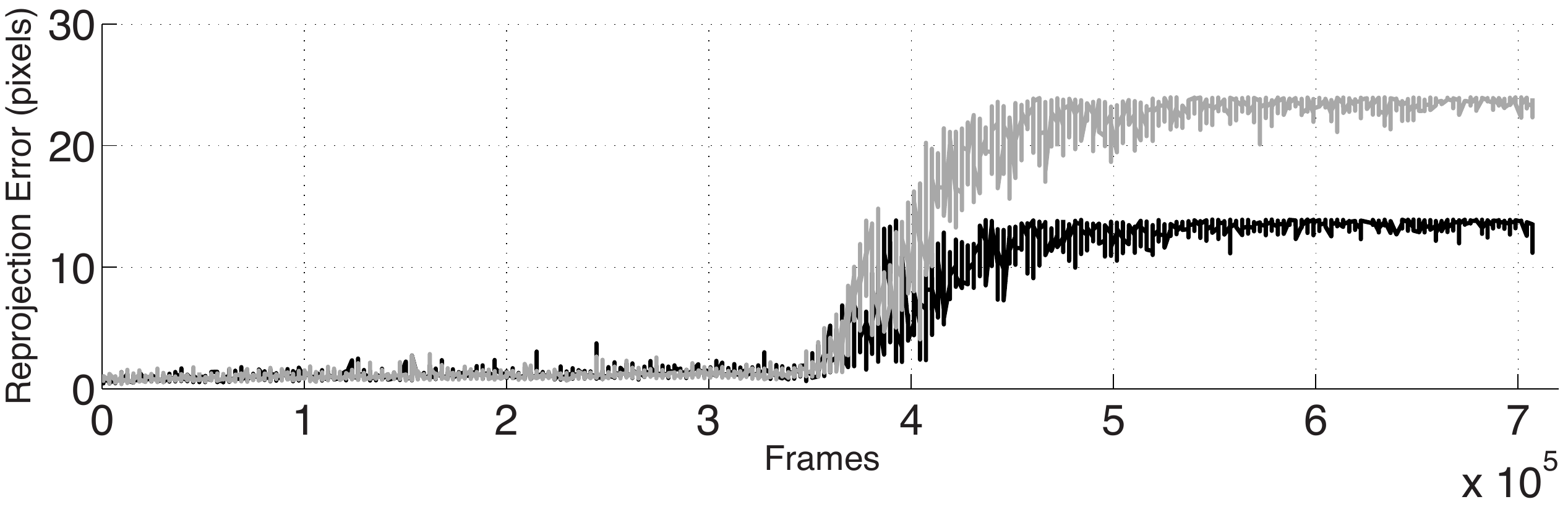}
\caption{Reprojection error over 8-hour operation for a sample keyframe without (light plot) and with (dark plot) \emph{proximity checking}.}
\label{error_repr}
\end{figure}

\begin{figure*}
\centering
\includegraphics[width=.95\textwidth]{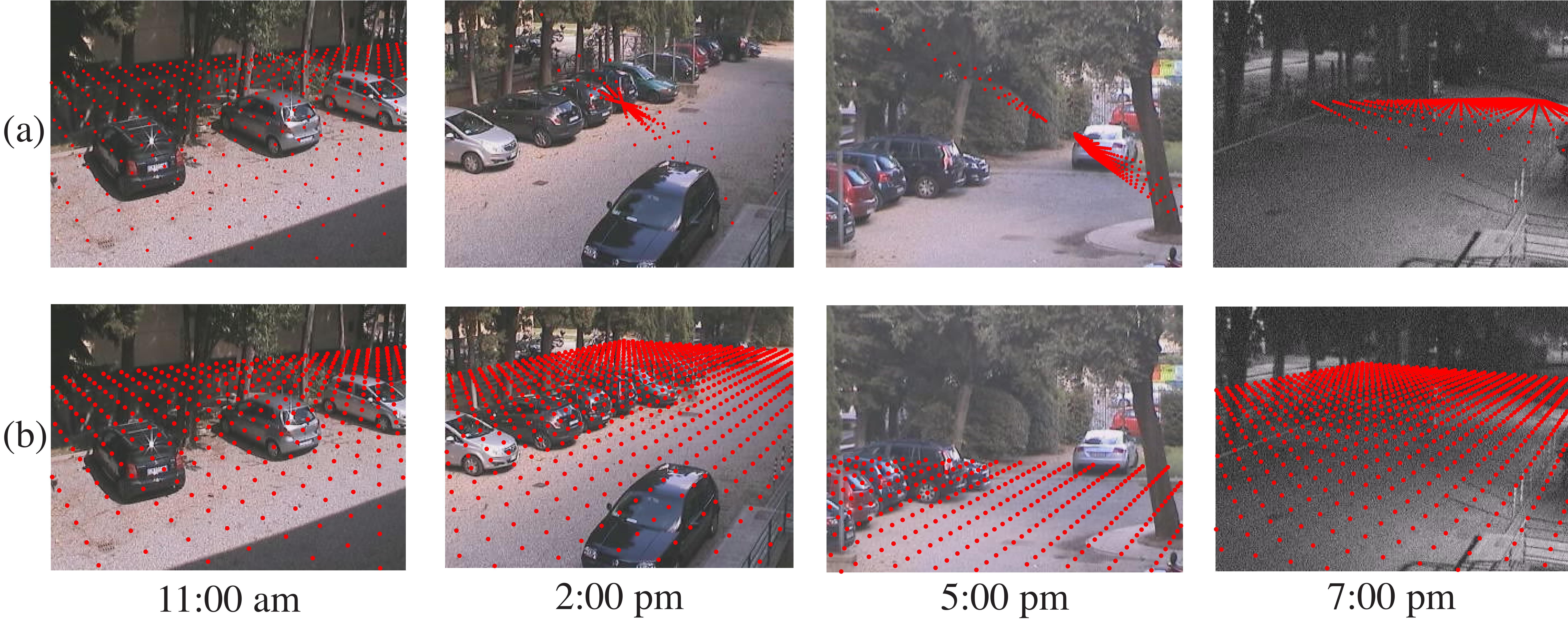}
\caption{Camera calibration without (a) and with scene map updating (b) at different time of the day.}
\label{fig:inliers2}
\end{figure*}

Our PTZ camera calibration keeps sufficiently stable over long periods of observation. Fig. \ref{error_repr} shows a typical plot of the reprojection error over 8-hour operation for a sample keyframe. Camera calibration at different time of the day without and with scene landmark updating is shown in Fig.~\ref{fig:inliers2}(a-b) for a few sample frames. It can be observed that with scene landmark updating, camera calibration (represented by the superimposed grid of points) is still accurate despite of the large illumination changes occurred in the scene.

\subsection{Multi-Target Tracking with PTZ cameras}

In the following, we summarize experiments on multi-target tracking in 3D world coordinates using our on-line PTZ camera calibration, and compare our method with a few methods that appeared in the literature on a standard PTZ video sequence. In our experiments targets were detected automatically using the detector in~\cite{dalal05}.

\subsubsection*{Influence of camera calibration}
To evaluate the impact of our PTZ calibration on tracking, we recorded a 8-hour sequence in a parking area during a working day and extracted three videos with one, two and three targets. This is a dynamic condition, with both smooth and abrupt scene changes. Multi-target tracking performance was evaluated according to both the CLEAR MOT~\cite{clear-MOT} and USC metrics~\cite{nevatia2011}. The CLEAR MOT metrics measures tracking accuracy (MOTA):

\begin{equation} \mbox{MOTA} = 1 - \frac{\sum_t (\mbox{FN}_t +
\mbox{FP}_t + \mbox{ID\_SW}_t )}{\sum_t \mbox{n}_t} 
\end{equation} 
and precision (MOTP):
\begin{equation} \mbox{MOTP} = \frac{\sum_{i,t} \mbox{VOC}_{i,t}}{\sum_t
\mbox{TP}_t} ,
\end{equation} 
where $\mbox{FN}_t$ and $\mbox{FP}_t$ are respectively the false negatives and positives, $\mbox{ID\_SW}_t$ are the identity switches, $\mbox{n}_t$ is the number of targets and $\mbox{VOC}_{i,t}$ is the VOC score of the $i$-th target at time $t$.
The USC metric reports the ratio of the trajectories that were successfully tracked for more than 80\% (MT), the ratio of mostly lost trajectories that were successfully tracked for less than 20\% (ML), the rest partially tracked (PT) and the average count of false alarms per frame (FAF). We measured the performance for the method with no scene map updating, with no \emph{proximity checking} and for the full method.

From Tab.~\ref{table:mota1} it is apparent that scene map updating has a major influence on the number of false negatives and false positives and therefore on the tracking accuracy. \emph{Proximity checking} has also a positive impact on the reduction of false positives and determines an average increase of the accuracy of about $10\%$.

\begin{table*}[tbh]
\centering
\resizebox{.9\textwidth}{!}{
\setlength{\tabcolsep}{0.4em}
\begin{tabular}{|l|c|c|c|c|c|c|c|c|c|c|}
\hline 	 
\multicolumn{1}{|l}{\textbf{Sequence and Method}} & \multicolumn{6}{|c}{\textbf{CLEAR MOT}} & \multicolumn{4}{|c|}{\textbf{USC Metric}} \\  \cline{1-11}
 	 & MOTA\%  & MOTP\% &  FN\% & FP\% & ID\_SW  & TR\_FR & MT\% & PT\% & ML\% & FAF \\ \cline{1-11}
\textbf{Seq.~\#1} (1 target) & & & & & & & & & & \\
\emph{Our method w/o map updating} &  -89.9 & 58.4 & 70.8 & 118.2 & 0 & 23 & 0.0 & 100.0 & 0.0 & 1.17  \\
\emph{Our method w/o proximity check} &  80.4 & 60.4 & 10.9 & 8.6 & 0 & \textbf{1} & 100.0 & 0.0 & 0.0 & 0.09  \\
\emph{Our method}                    &  \textbf{88.2} & \textbf{66.7} & \textbf{10.9} & \textbf{0.7} & \textbf{0} & 3 & \textbf{100.0} & \textbf{0.0} & \textbf{0.0} & \textbf{0.01}  \\
\hline
\textbf{Seq.~\#2} (2 target) & & & & & & & & & & \\
\emph{Our method w/o map updating}          &  -130.0 & 52.1 & 96.1 & 133.0 & 0 & 27 & 0.0 & 0.0 & 100.0 & 2.49  \\
\emph{Our method w/o proximity check}    &  70.4 & 61.5 & {25.7} & 3.6 & 0 & 10 & 50.0 & 50.0 & 0.0 & 0.07  \\
\emph{Our method}                            &  \textbf{78.8} & \textbf{64.2} & \textbf{19.4} & \textbf{1.6} & \textbf{0} & \textbf{8} & \textbf{50.0} & \textbf{50.0} & \textbf{0.0} & \textbf{0.03}  \\
\hline
\textbf{Seq.~\#3} (3 target) & & & & & & & & & & \\
\emph{Our method w/o map updating}          &  -51.5 & 59.4 & 81.9 & 69.1 & 0 & 20 & 0.0 & 66.7 & 33.3 & 2.06  \\
\emph{Our method w/o proximity check}     &  67.5 & \textbf{67.3} & 26.9 & 5.4 & 0 & 6 & 33.3 & 66.7 & 0.0 & 0.16  \\
\emph{Our method}                            &  \textbf{74.6} & 65.0 & \textbf{24.3} & \textbf{1.0} & \textbf{0} & \textbf{3} & \textbf{33.3} & \textbf{66.7} & \textbf{0.0} & \textbf{0.03}  \\
\hline
\end{tabular}
}
\caption{Multi-Target Tracking Performance in different settings: with one, two, three moving targets.}
\label{table:mota1}
\end{table*}

\subsubsection*{Influence of tracking in 3D world coordinates}

To analyze the effect of using 3D world coordinates we run our method in 2D image coordinates (not applying mapping in the 3D world plane). In this case, the target scale could not be evaluated directly and was estimated within a range from the scale at the previous frame. Tab.~\ref{table:mota2} reports the performance of our multi-target tracking performed in the two cases.

It can be observed that tracking in 3D world coordinates lowers the number of false positives and contributes to a sensible improvement in both accuracy and precision, with respect to tracking in the 2D image plane. This improvement is even greater as the number of targets increases since the tracker has to discriminate between them.

\begin{table*}[tbh]
\centering
\resizebox{.9\textwidth}{!}{
\setlength{\tabcolsep}{0.4em}
\begin{tabular}{|l|c|c|c|c|c|c|c|c|c|c|}
\hline 	 
\multicolumn{1}{|l}{\textbf{Sequence and Method}} & \multicolumn{6}{|c}{\textbf{CLEAR MOT}} & \multicolumn{4}{|c|}{\textbf{USC Metric}} \\  \cline{1-11}
 	 & MOTA\%  & MOTP\% &  FN\% & FP\% & ID\_SW  & TR\_FR & MT\% & PT\% & ML\% & FAF \\ \cline{1-11}
\textbf{Seq.~\#1} (1 target) & & & & & & & & & & \\
\emph{Our method in 2D}    &  79.9 & \textbf{70.6} & 15.1 & 4.9 & 0 & 3 & 100.0 & 0.0 & 0.0 & 0.05  \\
\emph{Our method in 3D}                    &  \textbf{88.2} & 66.7 & \textbf{10.9} & \textbf{0.7} & \textbf{0} & \textbf{3} & \textbf{100.0} & \textbf{0.0} & \textbf{0.0} & \textbf{0.01}  \\
\hline
\textbf{Seq.~\#2} (2 target) & & & & & & & & & & \\
\emph{Our method in 2D}    &  42.7 & 57.5 & 36.6 & 20.3 & 1 & 9 & 0.0 & 100.0 & 0.0 & 0.38  \\
\emph{Our method in 3D}                    &  \textbf{78.8} & \textbf{64.2} & \textbf{19.4} & \textbf{1.6} & \textbf{0} & \textbf{8} & \textbf{50.0} & \textbf{50.0} & \textbf{0.0} & \textbf{0.03}  \\
\hline
\textbf{Seq.~\#3} (3 target) & & & & & & & & & & \\
\emph{Our method in 2D}    &  59.5 & 62.5 & 31.8 & 8.5 & 0 & 7 & 0.0 & 100.0 & 0.0 & 0.25  \\
\emph{Our method in 3D}                    &  \textbf{74.6} & \textbf{65.0} & \textbf{24.3} & \textbf{1.0} & \textbf{0} & \textbf{3} & \textbf{33.3} & \textbf{66.7} & \textbf{0.0} & \textbf{0.03}  \\
\hline
\end{tabular}
}
\caption{Multi-Target Tracking Performance in 2D and 3D world coordinates.}
\label{table:mota2}
\end{table*}

\begin{table*}[t]
\centering
\resizebox{.9\textwidth}{!}{
\setlength{\tabcolsep}{0.4em}
\begin{tabular}{|l|c|c|c|c|c|c|c|c|c|c|}
\hline 	 
\multicolumn{1}{|l}{\textbf{Sequence and Method}} & \multicolumn{6}{|c}{\textbf{CLEAR MOT}} & \multicolumn{4}{|c|}{\textbf{USC Metric}} \\  \cline{1-11}
 	 & MOTA\%  & MOTP\% &  FN\% & FP\% & ID\_SW  & TR\_FR & MT\% & PT\% & ML\% & FAF \\ \cline{1-11}
 \textbf{UBC Hockey} (Okuma's detector) & & & & & & & & & & \\
Okuma~\cite{kenjieccv2004}    &  67.8 &   51.0 & 31.3 & 0.0 & 11 & 0 & -- & -- & -- & -- \\
\emph{Our method in 2D}   & 67.9 &  \textbf{62.3} & 8.8 & 23.2 & 0 & 1 & 91.7 & 8.3 & 0 & 2.47 \\
\emph{Our method in 3D}   & \textbf{90.3} &  60.4 & \textbf{6.5} & \textbf{3.1} & \textbf{0} & \textbf{1} & \textbf{91.7}  & \textbf{8.3} & \textbf{0} & \textbf{0.35}\\
\hline
 \textbf{UBC Hockey} (ISM~\cite{bleibe-ism-detector08} detector)  & & & & & & & & & & \\
Breitenstein~\cite{tpami-Breitenstein}  & 76.5  &  57.0 & 22.3 & 1.2 & 0 & -- & -- & -- & -- & -- \\
Brendel~\cite{Brendel:cvpr11} &  79.7 & 60.0  & 19.5 & \textbf{1.1} & 0 & -- & -- & -- & -- & -- \\
\emph{Our method in 2D} & 72.6 & 61.0  & 18.7 & 8.6 & 0 & 1 & 58.3  & 33.3 & 8.3 & 0.93  \\
\emph{Our method in 3D}   & \textbf{83.6} &  \textbf{63.8} & \textbf{14.5} & 1.9 & \textbf{0} & \textbf{0} & \textbf{75} & \textbf{16.7} & \textbf{8.3} & \textbf{0.21} \\
\hline
\end{tabular}
}
\caption{Multi-Target Tracking performance on UBC Hockey dataset.}
\label{table:mota3}
\end{table*}

We compared our calibration and tracking against the results reported by a few authors, namely~\cite{kenjieccv2004},~\cite{tpami-Breitenstein} and~\cite{Brendel:cvpr11}, on the \emph{UBC Hockey} sequence~\cite{kenjieccv2004}. This is the only publicly available dataset recorded from a PTZ camera. It is very short and includes frames of a hockey game. All these authors performed tracking in the 2D image plane. For the sake of completeness we have compared both the 2D and 3D version of our tracking method. The scene map was obtained by uniformly sampling the video sequence every ten frames so to have a full coverage of the scene. 
For a fair comparison, in a first experiment we compared our method against~\cite{kenjieccv2004} using the original detections provided by Okuma.  In a second experiment we compared with~\cite{tpami-Breitenstein} and~\cite{Brendel:cvpr11} using the ISM detector~\cite{bleibe-ism-detector08}.
The results are reported in Tab.~\ref{table:mota3}. As it is possible to observe, in the first experiment our calibration and tracking in 2D coordinates obtains comparable performance as~\cite{kenjieccv2004},  while tracking in 3D world coordinates has significantly superior performance.  In the second experiment, we observed that the ISM detector fails to detect a target in the entire sequence and determines a large number of false negatives in all the methods. Notwithstanding calibration and tracking in 3D coordinates still reports some improvement in performance with respect to the 2D solutions.

\subsection{Operational Constraints and Computational requirements}

We analyzed the operational constraints and computational requirements of our solution using a SONY SNC-RZ30P PTZ camera and Intel Xeon Dual Quad-Core at 2.8GHz and 4GB of memory, with no GPU processing. 
From Tab.~\ref{tab:timings}  we can see that we perform real-time calibration and tracking (in 3D world coordinates) at 12 fps. 
The current implementation of the method exploits multiple cores and was developed in C/\CC. Frame grabbing, camera calibration and scene map updating are performed in one thread, detection and tracking are performed in a separate thread.

\begin{table}[h]
\centering
\resizebox{.9\columnwidth}{!}{
\begin{tabular}{|l|r|r|}
  \hline
  Component & Time & fps\\
    \hline
  Camera Pose Estimation & 88 ms & 11 \\
  Scene Map Update & 5 ms & 200 \\
  Detection & 43 ms & 23 \\
  Tracking & 35 ms & 28  \\
  \hline \hline
  Total (Sequential) & 171 ms & 5 \\
  \hline \hline
  \textbf{Total (Parallel}) & 83 ms & \textbf{(x2.4) 12} \\
  \hline
\end{tabular}
}
\caption{Computational requirements per processing module on a Intel Xeon Dual Quad-Core at 2.8GHz.}
\label{tab:timings}
\end{table}

\section{Conclusions}
\label{sec:conclusion}

In this paper, we have presented an effective solution for on-line PTZ camera calibration that supports real-time multiple target tracking with high and stable degree of accuracy. Calibration is performed by exploiting the information in the current frame and has proven to be robust to camera motion, changes of the environment due to illumination or moving objects and scales beyond thousands of landmarks. 
The method directly derives the relationship between the position of a target in the 3D world plane and the corresponding scale and position in the 2D image. This allows real-time tracking of multiple targets with high and stable degree of accuracy even at far distances and any zooming level.


\bibliographystyle{spphys}       

\end{document}